\documentclass[
]{ceurart}

\usepackage[utf8]{inputenc}
\usepackage{listings}
\usepackage{svg}
\usepackage{placeins}
\usepackage{placeins}
\usepackage{hyperref}
\usepackage[most]{tcolorbox}
\usepackage{makecell} 
\begin{document}

\copyrightyear{2026}
\copyrightclause{Copyright for this paper by its authors.
  Use permitted under Creative Commons License Attribution 4.0
  International (CC BY 4.0).}

\conference{CLEF 2026 Working Notes, 21 -- 24 September 2026, Jena, Germany}

\title{DS@GT ARC at CheckThat! 2026: LLM-Based Trace Ranking and Grouped Reward Modeling for Multilingual Numerical Claim Verification}

\title[mode=sub]{CheckThat! at CLEF 2026}

\tnotemark[1]

\author[1]{Sagnik Sinha}[%
orcid=0009-0002-7290-7317,
email=ssinha348@gatech.edu,
url=https://github.com/sagniksinha,
]
\cormark[1]
\fnmark[1]

\author[1]{Shreyas Shrestha}[%
orcid=0009-0005-2061-3546,
email=shreyas.shrestha@gatech.edu,
url=https://github.com/shreyas-shrestha,
]
\fnmark[1]
\address[1]{Georgia Institute of Technology, North Ave, Atlanta, GA 30332, United States}

\cortext[1]{Corresponding author.}
\fntext[1]{These authors contributed equally.}

\begin{abstract}
  Automated verification of numerical claims is a challenging problem, as it requires 
both language understanding and quantitative reasoning. This paper describes our 
system for CLEF 2026 CheckThat! Task 2, which focuses on ranking reasoning traces 
generated by large language models (LLMs) and predicting a final verdict for 
numerical claims in English and Arabic. We explore two approaches. The first 
approach fine-tunes an LLM-based verifier using LoRA to score each reasoning trace 
independently as a binary classification problem, and selects the final verdict using 
Best-of-N selection. We further experiment with adaptive sub-claim decomposition to 
break complex claims into simpler parts before verification. The second approach uses 
a lightweight TF-IDF reward model with handcrafted numeric and temporal overlap 
features to score traces, and aggregates scores by verdict group to determine the 
final prediction. For Arabic, we compare a general multilingual model against 
AraBERT, a language-specific model pretrained on Arabic text. Our results show that 
the LLM-based approach outperforms the lightweight reward model on most metrics, 
particularly Recall@5, while the reward-based approach shows stronger performance on 
the Conflicting class. Sub-claim decomposition did not improve performance, 
suggesting that claim splitting introduces noise rather than aiding reasoning. For 
Arabic, AraBERT outperforms the multilingual baseline across most metrics.
\end{abstract}


\maketitle

\section{Introduction}

The fast spread of misinformation has become one of the most pressing challenges of the current digital age. With billions of people consuming news through social media platforms, false or misleading information can reach wide audiences within minutes, far outpacing any manual attempt to correct it \cite{aimeur2023fake}. Since many adults rely on social media as their main source of news \cite{pewresearch2023}, the risk of misinformation spreading further is greater than ever. Traditional fact-checking organizations, such as PolitiFact and Snopes, rely on trained journalists to manually verify claims, a process that is both time-consuming and impossible to scale against the sheer volume of content circulating online \cite{shaar2021assisting}. This has made the development of automated fact-checking systems a critical research priority.

This problem has become larger in scale and more complex with the rise of large language models (LLMs) and AI-generated online content. Although these tools are highly capable, they also make it easier to produce convincing but factually incorrect text at scale. Thus, developing robust and reliable automated fact-verification systems is crucial for maintaining trust in publicly available information.

A particularly challenging part of this problem is the need to verify numerical claims. Numerical claims are statements that include specific quantities, statistics, dates, or other temporal expressions. Such claims are especially dangerous in misinformation contexts due to the well-documented \textit{numeric-truth effect}, a cognitive bias in which the presence of numbers makes a statement appear more credible, regardless of its accuracy \cite{sagara2009numeric}. Verifying numerical claims involves not only understanding language, but also performing fine-grained quantitative reasoning. This is a capability that remains difficult for current automated systems \cite{runewicz2025fraunhofer, heil2025dsgt}.

The CheckThat! Lab at CLEF has long been addressing this challenge through its annual shared tasks and competitions. These competitions benchmark systems across different components of the fact-checking pipeline \cite{nakov2018checkthat, barron2020checkthat, barron2024checkthat}. In the 2025 edition, Task 3 focused specifically on the automated verification of numerical and temporal claims \cite{alam2025checkthat}. Using the QuanTemp dataset — a large scale benchmark drawn from 45 fact-checking organizations worldwide, comprising over 15,000 annotated claims and an evidence corpus of over 423,000 snippets \cite{venktesh2024quantemp} — participants were tasked with classifying each claim as \textit{True}, \textit{False}, or \textit{Conflicting} based on retrieved evidence. The task represented an important first step toward dedicated benchmarking of numerical fact-checking, but it evaluated systems primarily on classification accuracy without examining the quality of the reasoning that led to each verdict.

With the CLEF 2025 foundation in place, CLEF 2026 CheckThat! Task 2 introduces a substantially new and more ambitious subset of the problem. Rather than simply predicting a verdict, this year's task introduces a test-time scaling framework that evaluates the ranking of reasoning traces itself \cite{clef-checkthat:2026:task2}. Given a claim, associated evidence, and multiple reasoning traces generated by an LLM at varying temperature settings to encourage diversity, participants must train a verifier model that ranks these reasoning traces by their utility in arriving at the correct verdict and then derives a final verdict from the top-ranked traces \cite{liang2026gr2}.

The CLEF 2026 setup evaluates both the correctness of a system's verdict and the quality of its underlying reasoning. This is an important step forward because it assesses whether automated systems can evaluate the reasoning behind a verdict, rather than simply classifying it. Another important aspect of the 2026 task is its multilingual dataset. While the 2025 task was conducted in English only, this year's task extends to Spanish and Arabic, drawing on 2,808 Spanish and 3,260 Arabic claims from the previous edition alongside the English QuanTemp data. The scope of this paper is limited to English and Arabic only. Systems are evaluated using Macro-averaged F1 for verdict accuracy and Recall@k and MRR@k for reasoning trace ranking quality, with the final score averaging performance across both dimensions.

In this paper, we describe our system for CLEF 2026 CheckThat! Task 2. We address the following three Research Questions (RQ):
\begin{itemize}
\item \textbf{RQ1:} Can a single multilingual model handle all languages, i.e. English and Arabic? Or will performance improve with language-specific models?
\item \textbf{RQ2:} Does breaking a claim into smaller sub-claims improve reasoning trace ranking performance?
\item \textbf{RQ3:} Can a grouped reward-based ranking model centered on trace-level analysis improve final verdict prediction?
\end{itemize}
The remainder of the paper is organized into the following sections: Section 2 reviews related work; Section 3 describes our methodology; Section 4 presents results; Section 5 discusses future work; and Section 6 concludes the paper.

The competition uses Recall@k and MRR@k, with k=5, to measure reasoning-trace ranking quality. For claim verification, we report macro-averaged F1 and class-wise F1 scores. The final score is obtained by averaging the macro-F1 and Recall@5 scores.

\section{Related Work}
Numerical claim verification is a distinct problem in natural language processing because it requires both accurate semantic interpretation and explicit temporal and numerical grounding. Accordingly, prior CheckThat! submissions have often used multi-stage architectures that separate evidence selection, evidence scoring, reasoning, and final verdict prediction. Fraunhofer SIT used a three-stage architecture with dense evidence retrieval, contrastive re-ranking, and evidence pooling for the final verdict \cite{runewicz2025fraunhofer}. SINAI-UGPLN used a multi-stage meta ensemble framework, integrating numeric and lexical features alongside their transformer-based classifiers. Earlier DS@GT groups examined architectural differences within retrieval, context construction, as well as tokenization before producing a final verdict \cite{heil2025dsgt}. 

Our work builds on these architectures by shifting supervision to the reasoning-trace level. Instead of assessing only the claim-level input, we assign each trace a binary utility label based on whether or not its verdict matches the gold claim. In Approach I, we pair this trace-level supervision with sub-claim decomposition, inspired by prior work where similar claim decomposition methods and adaptive claim decomposition methods \cite{wanner2024closer} have been shown to boost performance of LLM-based solutions for tasks involving reasoning.

Furthermore, previous CheckThat! teams have shown numerical claim verification can benefit through the implementation of tokenization, and numeric/lexical features \cite{heil2025dsgt}. In Approach II, we build a feature-driven model by using word-level and character-level TF-IDF representations with numeric and temporal overlap and mismatch also tracked. We were also inspired by Fraunhofer SIT's approach to use multi-instance evidence pooling to accumulate support across many smaller scored units, as opposed to a single top ranked item \cite{runewicz2025fraunhofer}. This motivates our grouped trace-ranking architecture which measures support across multiple traces assigned to each verdict type (True, False, and Conflicting) before deciding the final verdict.

For the claim decomposition approach, we adopt the framework introduced by Wanner et al. (2024)  \cite{wanner2024closer}, who investigate LLM-based claim decomposition. Their study demonstrates that decomposing complex claims into atomic sub-claims improves the reliability of downstream factual verification. Importantly, they show that the effectiveness of verification is strongly dependent on the quality of the decomposition, thereby motivating the use of prompt-based decomposition as a preprocessing step prior to claim verification.

\section{Methodology}
We explored two complementary approaches for CheckThat! Task 2. Both approaches shared the same core supervision principle, as each reasoning trace was assigned a binary utility label indicating whether or not its verdict matched the gold claim label. However, the two approaches differed in how traces were sampled and represented, and how the final claim-level verdict was constructed. Approach I focuses on fine-tuning LLM models using LoRA, integrating sub-claim decomposition, and also incorporates an AraBERT encoder stream for Arabic text. Approach II estimates trace utility using a lightweight TF-IDF-based reward model, and then obtains the final verdict using a grouped trace utility ranking across the True, False, and Conflicting verdict groups.

\subsection{Approach I: LLM-Based Architectures}
\label{subsec:approach1}
Approach I frames reasoning-trace ranking as a binary classification problem. We also experimented with claim decomposition and focal loss. Because focal loss reduced performance, we omit it from the main comparison.
\subsubsection{Designing the Classification Problem}
The training data is passed through a data pipeline where each training example combines a claim with one of its reasoning traces. For each claim, up to six traces are sampled, i.e. roughly two that match the label and the rest drawn from the opposing verdict types (True, False or Conflicting), with a small allowance for unknown verdicts to avoid skewing the distribution. The training target variable is set as a binary variable. Every reasoning trace receives label 1 if its verdict matches the gold label of the claim, and 0 otherwise. The model input concatenates the claim text and the trace verdict. This formulation creates a setup where the modeling becomes a binary classification task. Model fine-tuning was performed using LoRA on the training dataset.
\subsubsection{Sub-Claim Decomposition}
We also experimented with sub-claim decomposition, where each claim was decomposed to smaller sub-claims. A common first step, particularly relevant in real-world scenarios where the given claims are naturally occurring, often lengthy sentences, is to decompose the claim into smaller sub-claims, making it easier to verify each sub-claim separately \cite{min-etal-2023-factscore, kamoi-etal-2023-wice, mitra-etal-2025-factlens}.
\begin{figure}[h]
    \centering
    \includegraphics[width=0.8\textwidth]{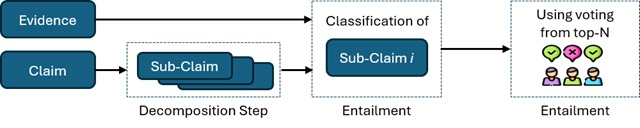}
    \caption{Claim-decomposition: The architecture consists of the following steps - (i) Decomposition: the original claim is split into individual sub-claims. (ii) Entailment: each sub-claim is classified individually. (iii) Aggregation: performed using voting.}
    \label{fig:your_label}
\end{figure}
\FloatBarrier
\vspace{-1.5em}
To better handle complex claims containing multiple factual assertions, we introduce an LLM-based claim decomposition step. Before verification, each claim is passed to an instruction-tuned Llama-3.2-3B-Instruct model, which is prompted to decompose the original claim into at most two independent and concise sub-claims. If a claim cannot be meaningfully split, the model returns the original claim unchanged. During training, each generated sub-claim is treated as an additional training instance while preserving the original verdict and justification. During inference, the verifier scores both the original claim and its decomposed sub-claims, and the final verification score is obtained by averaging (or weighted averaging) these scores. This decomposition enables the verifier to evaluate individual factual components separately, improving robustness for multi-faceted claims.
\begin{tcolorbox}[
title=LLM Prompt for Claim Decomposition,
colback=gray!5,
colframe=black,
fonttitle=\bfseries
]
\ttfamily
Decompose the following claim into at most 2 independent, concise
sub-claims.

Return ONLY a numbered list (1. ..., 2. ...).
If the claim cannot be meaningfully split, return it unchanged as item 1.

Claim: \{claim\_text\}

Sub-claims:
\end{tcolorbox}
\subsubsection{Language Specific Approach}
To address the multilingual nature of the task, we experimented with both a single cross-lingual model and a language-specific model for Arabic datasets. For the cross-lingual setting, the same model trained on English data was applied directly to Arabic claims without any language-specific adaptation. For the language-specific setting, we used AraBERT \cite{antoun2020arabert}, a BERT-based model pre-trained on large Arabic corpora, as the backbone encoder for Arabic claims. AraBERT has demonstrated state-of-the-art performance across a range of Arabic NLP benchmarks, outperforming multilingual BERT in Arabic-specific tasks. We compared these two settings to examine whether a dedicated Arabic encoder provides meaningful gains over a general cross-lingual model for the trace ranking and verdict prediction subtasks.

\subsection{Approach II: Grouped Reward-Based Trace Ranking}
\label{subsec:approach2}
\begin{figure}[h]
    \centering
    \includegraphics[width=0.8\textwidth]{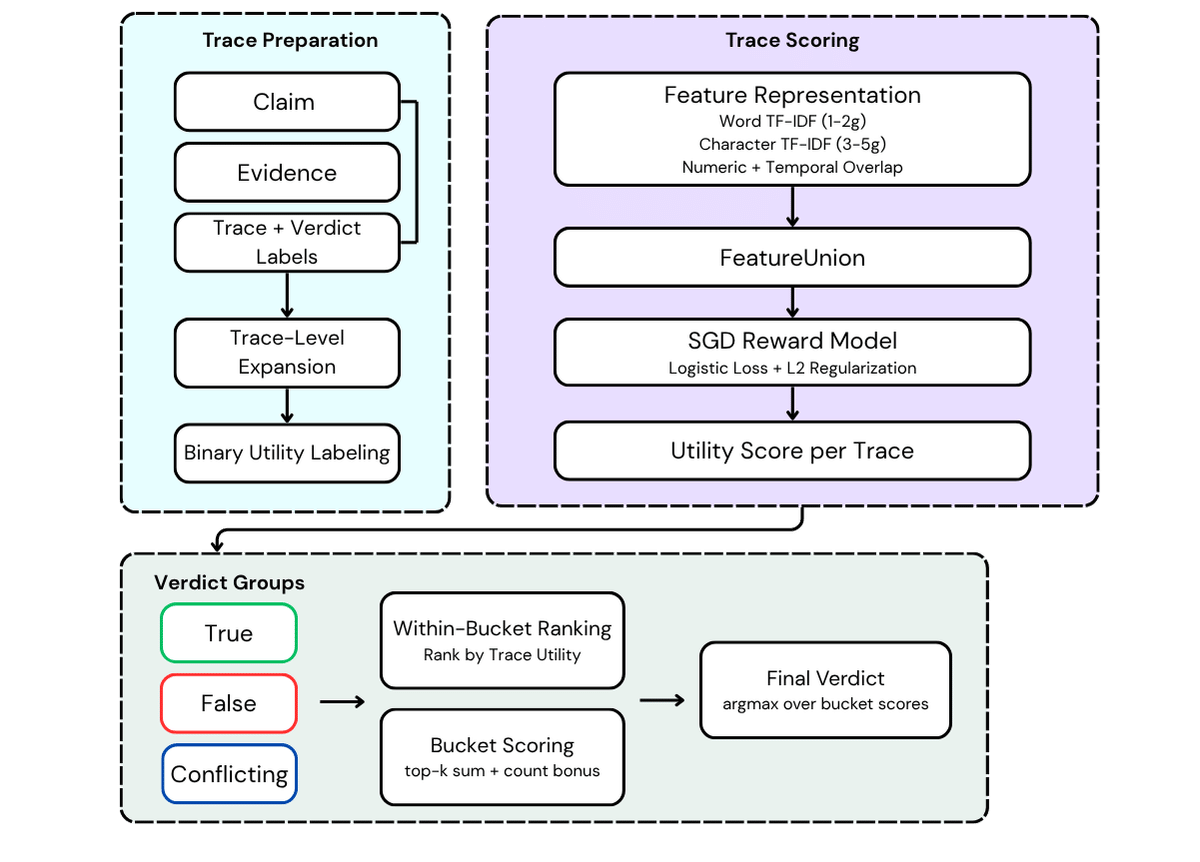}
    \caption{Overview of Approach II. The system converts each claim into trace-level examples, assigns binary utility labels, extracts numeric and temporal features, and scores each trace using a lightweight reward model. Trace scores are grouped by verdict label, ranked within each verdict group, and aggregated to produce the final verdict.}
    \label{fig:my_label}
\end{figure}

Approach II uses a grouped reward trace ranking architecture designed to examine whether an interpretable, lightweight verifier can provide a signal of trace utility without the need for a fine-tuned neural re-ranker. The pipeline consists of four main stages. First, each claim is restructured into trace-level training rows. Second, each trace is assigned a binary utility label dependent on whether its verdict matches the gold claim label. Third, a lightweight TF-IDF-based reward model is used to estimate trace utility. Finally, trace scores are grouped by verdict label (True, False, and Conflicting), then aggregated within each verdict group, and used to create the final verdict. Figure~\ref{fig:my_label} provides an overview of the full Approach II pipeline.
\subsubsection{Trace-Level Restructuring}
Each original claim is associated with multiple reasoning traces, corresponding verdict labels, and evidence. We convert this claim-level structure into a trace-level structure for our training set to better align with the task. Each trace is transformed into one row containing the claim text, evidence text, trace text, trace verdict, and gold label of the claim. This allows the model to evaluate one trace-verdict pair at a time, rather than the entire claim and all of its associated traces as a single training instance.
\subsubsection{Utility Supervision}
We label each trace with a binary utility value. A trace receives a utility value of 1 if the verdict associated with that trace matches the gold label of the claim, and a value of 0 otherwise. This procedure allows the model to learn whether a trace-verdict pairing should be trusted or not for the current claim. Through this methodology, the task is reframed around trace utility rather than direct claim classification.
\subsubsection{Feature Representation}
For each trace row, we construct a combined representation using the claim, evidence, trace verdict, trace text, and a custom feature block. The feature block utilizes regular expression extraction on numeric values, years, and months, in order to identify overlap and mismatches across the claim, evidence, and trace. This includes finding the number of numeric tokens in the claim, evidence, and trace, the number of shared numeric tokens across pairs of those texts, and the number of claim-level numeric tokens that are missing from the trace. We apply the same overlap methodology on year and month tokens. We also include indicators for percentages and currency markers in the feature block. The resulting total representation captures whether or not a trace preserves the numeric and temporal values that appear in the claim and evidence. This is critical for numerical claim verification because a trace may appear to be plausible, but in actuality may omit, alter, or add quantities which ultimately determine the final verdict.
\subsubsection{Reward Model}
We used a lightweight reward model to test whether lexical features and numeric and temporal overlap signals could provide useful estimates of trace utility without the burden of fine-tuning a transformer-based neural re-ranker. Specifically, TF-IDF was selected because it is efficient, reproducible, and simple to combine with custom features. This allowed Approach II to be a useful and interpretable comparison point against the LLM-based verifier used in Approach I. 

Approach II used a FeatureUnion over a word-level TF-IDF (Term Frequency-Inverse Document Frequency) vectorizer with n-gram range = (1,2) and a character-level TF-IDF vectorizer with n-gram range = (3,5). These vectorizers were then applied to a combined representation which contained the claim, evidence, trace text, trace verdict, and a custom numeric and temporal feature block. This combined representation was passed to an SGD classifier with logistic loss, L2 regularization, and weight averaging. Balanced class weights were computed over the binary utility labels mentioned above for training. Ultimately, the reward model was used to then compute trace-level utility scores for grouped verdict aggregation.
\subsubsection{Grouped Verdict Aggregation}
After the reward model scored all traces for a given claim, the system grouped these traces by their verdict labels of True, False, and Conflicting. Within these verdict groups, traces were then sorted by their utility score generated by the reward model. The top-k trace utility scores and a count bonus based on the number of traces contributing to the top-k scores were combined to establish a total score for each verdict label. Critically, to ensure a verdict without any supporting traces was not selected, empty verdict groups were assigned a score of \(-1.0\). The hyperparameters were tuned on the validation set by testing various top-k group sizes \{1, 2, 3, 5\} and count bonus values \{0.00, 0.02, 0.03, 0.05\}. The best validation setting used \(k=5\) and a count bonus of \(0.05\), selected using the combined score \(0.5 \times (\text{Macro-F1} + \text{Recall@5})\). The final verdict was then chosen based on the verdict label with the highest group score. This grouped aggregation methodology was designed to handle verdict support being spread across multiple traces rather than being concentrated into one. This is particularly relevant for the Conflicting class, where it is likely that evidence is split across many competing verdicts. This may explain why Approach II has a stronger Conflicting F1 score, even though it has a lower Recall@5 compared to Approach I.
\section{Results}
All results reported below are on the test set. Experiments were run separately for English and Arabic, and results for each language are presented in the following subsections.
\subsection{English Language}
Table~\ref{tab:English_results} summarizes the performance of all approaches on the English test set. Approach I, which uses an LLM-based architecture, outperforms Approach II overall, achieving significantly higher Recall@5. However, Approach II performs better on the Conflicting class (18.02 vs. 12.74), suggesting that the grouped reward-based ranking is better at capturing harder, ambiguous cases. Approach I struggles most on the Conflicting class, which is consistent with the general difficulty of this category across all approaches.

For Approach I, several LLM backbones were evaluated, including Llama-3.2-1B, ModernBERT-large, and Qwen2.5-Math-7B. Performance was close across all three, with Qwen2.5-Math-7B achieving the best results. The models were compared using a composite score defined as the average of Macro-F1 and Recall@5. Qwen2.5-Math-7B obtained a score of 41.88, followed by Llama-3.2-1B with 41.42, and ModernBERT-large with 40.78. The improvement of Qwen2.5-Math-7B is likely due to its pretraining on mathematical reasoning tasks, making it better suited for the task of numerical claim verification \cite{yang2024qwen25mathtechnicalreportmathematical}. The final selection was Qwen2.5-Math-7B.

Implementation Details. The final model builds on the Qwen2.5-Math-7B backbone, loaded in bfloat16 precision with all base-model parameters frozen. Parameter-efficient fine-tuning is performed using LoRA, with rank $r = 8$, scaling factor $\alpha = 16$, and dropout of $0.1$, with adapters injected into all attention and feed-forward projection layers, namely \texttt{q\_proj}, \texttt{k\_proj}, \texttt{v\_proj}, \texttt{o\_proj}, \texttt{gate\_proj}, \texttt{up\_proj}, and \texttt{down\_proj}. A single linear classification head is attached on top of the final hidden representation to produce binary predictions. Input sequences are tokenized with a maximum length of $1{,}024$ tokens, right-padded to this length. The model is trained for $20$ epochs with a batch size of $10$, using the AdamW optimizer (learning rate $2 \times 10^{-5}$, $\epsilon = 1 \times 10^{-8}$, weight decay $0.01$) and a cosine learning-rate schedule with linear warmup over the first $5\%$ of total training steps, with gradients clipped to a maximum norm of $1.0$ and optimization performed using standard cross-entropy loss. The dataset is split into training and validation subsets using an $80/20$ stratified split with a fixed random seed ($\texttt{random\_state} = 42$) to preserve class balance across splits. A model checkpoint is saved at the end of every epoch, and the checkpoint corresponding to the epoch with the highest validation accuracy is selected as the final model used for test-time inference.

Adding sub-claim decomposition to Approach I led to a drop in performance across most metrics, including Macro-F1 (55.79 vs. 58.98) and Recall@5 (23.57 vs 24.77). This suggests that breaking claims into sub-claims introduces noise into the pipeline rather than helping the model reason more precisely.
\begin{table}[h]
    \centering
    \caption{Results of approaches on the English test set.}
    \label{tab:English_results}
    \begin{tabular}{lcccccc}
        \hline
        \textbf{Approach} & \textbf{Score} & \textbf{Macro Avg F1} & \textbf{Recall@5} & \textbf{True F1} & \textbf{Conflicting F1} & \textbf{False F1} \\
        \hline
        Approach I & 41.88 & 58.98 & 24.77 & 72.42 & 12.74 & 88.77 \\
        \makecell[l]{Approach I \\ + Sub-Claims} & 39.68 & 55.79 & 23.57 & 64.68 & 13.33 & 89.34 \\
        Approach II & 40.50 & 58.78 & 22.21 & 70.44 & 18.02 & 87.88 \\
        \hline
    \end{tabular}
\end{table}
\FloatBarrier
Although Approach II underperforms Approach I overall, its higher Conflicting F1 suggests that grouped aggregation may provide a useful signal for ambiguous cases where evidence support is distributed across competing verdicts. Because conflicting claims often contain weaker support from multiple traces, rather than being concentrated in one trace, grouped aggregation better preserves repeated support for a verdict label across traces. This may explain why Approach II has a much higher Conflicting F1 score compared to Approach I, despite having a much lower overall Recall@5.
\subsection{Arabic Language}
Table~\ref{tab:Arabic_results} shows the results on the Arabic test set. Several LLM variants were tested on Arabic data; here we focus on comparing BERT against AraBERT as the key contrast. AraBERT outperforms BERT on most metrics, including Macro-F1 (84.85 vs. 82.32) and Recall@5 (33.41 vs. 29.54). This is expected, as AraBERT is specifically trained on Arabic text, which gives it stronger language understanding compared to a general multilingual model. Both models were fine-tuned using LoRA. In the test set no conflicting label is present, hence Conflicting F1 is not reported.
\begin{table}[h]
    \centering
    \caption{Results of approaches on the Arabic test set.}
    \label{tab:Arabic_results}
    \begin{tabular}{lccccccc}
        \hline
        \textbf{Approach} & \textbf{Score} & \textbf{Macro Avg F1} & \textbf{Recall@5} & \textbf{True F1} & \textbf{Conflicting F1} & \textbf{False F1} \\
        \hline
        AraBERT & 59.13 & 84.85 & 33.41 & 82.19 & - & 87.52 \\
        BERT & 55.93 & 82.32 & 29.54 & 83.25 & - & 83.38  \\
        \hline
    \end{tabular}
\end{table}

The final model for the Arabic setting builds on the AraBERT backbone (\texttt{aubmindlab/bert-base-arabertv02}), loaded in bfloat16 precision as an encoder-only model with all base-model parameters frozen. Parameter-efficient fine-tuning is performed using LoRA, with rank $r = 8$, scaling factor $\alpha = 16$, and dropout of $0.1$, with adapters injected into the encoder's query, key, value, and dense projection layers. A single linear classification head is attached on top of the pooled hidden representation to produce binary predictions. Input sequences are tokenized with a maximum length of $512$ tokens (selected based on the 95th percentile of the training-set token-length distribution), right-padded to this length using the tokenizer's unknown token as the padding token. The model is trained for $20$ epochs with a batch size of $10$, using the AdamW optimizer (learning rate $2 \times 10^{-5}$, $\epsilon = 1 \times 10^{-8}$, weight decay $0.01$) and a cosine learning-rate schedule with linear warmup over the first $5\%$ of total training steps, with gradients clipped to a maximum norm of $1.0$ and optimization performed using standard cross-entropy loss. The dataset is split into training and validation subsets using an $80/20$ stratified split with a fixed random seed ($\texttt{random\_state} = 42$) to preserve class balance across splits. A model checkpoint is saved at the end of every epoch, and the checkpoint corresponding to the epoch with the highest validation accuracy is selected as the final model used for test-time inference.

\section{Future Work}
Several promising directions remain for future work. One key direction is to enhance Approach II by training it on larger and more diverse external datasets. Since Approach II was trained from scratch on the provided training data, incorporating additional data could improve its ability to generalize and capture a wider range of reasoning patterns and numerical expressions.

Another promising direction is to combine both approaches into an ensemble. The 
results show that Approach I and Approach II have complementary strengths — Approach 
I performs better overall and on Recall@5, while Approach II is stronger on the 
Conflicting class. An ensemble that combines the scores or predictions of both 
approaches could help overcome the individual limitations of each, potentially 
yielding stronger and more balanced performance across all verdict classes.

More advanced claim decomposition strategies beyond prompt-based LLM decomposition could be explored, including learned or retrieval-guided decomposition methods.
Such approaches may reduce decomposition noise and improve robustness when handling complex multi-faceted numerical claims.

Additionally, current systems produce a verdict but limited explanations are offered by the model. Future work could focus on generating human-readable justifications alongside the verdict, improving interpretability for end users.

\section{Conclusions}
This paper presented two approaches for reasoning trace ranking and verdict prediction 
for numerical claims in English and Arabic. Approach I, an LLM-based verifier 
fine-tuned with LoRA, achieved stronger overall performance, while Approach II, a 
lightweight TF-IDF reward model, showed a clear advantage on the harder Conflicting 
class. Sub-claim decomposition did not improve performance, and AraBERT outperformed 
multilingual BERT for Arabic. Across all settings, the Conflicting class remained 
the most difficult category, pointing to the broader challenge of verifying ambiguous 
numerical claims. These results suggest that the LLM-based approach is more effective overall, while grouped reward modeling may provide a complementary signal for ambiguous numerical claims.

\section*{Acknowledgments}

We thank the Data Science at Georgia Tech (DS@GT) CLEF competition group for their support.
This research was supported in part through research cyberinfrastructure resources and services provided by the Partnership for an Advanced Computing Environment (PACE) at the Georgia Institute of Technology, Atlanta, Georgia, USA \cite{PACE}. 

\section*{Declaration on Generative AI}
 During the preparation of this work, the authors used ChatGPT, Gemini for: Grammar and spelling check, Drafting content. After using these tools, the authors reviewed and edited the content as needed and take full responsibility for the publication’s content.

\bibliography{ceur}

\section{Online Resources}
The code and implementation for this work are available on:
\begin{itemize}
    \item GitHub repository: \href{https://github.com/dsgt-arc/clef2026-checkthat-task2/}{https://github.com/dsgt-arc/clef2026-checkthat-task2/}
\end{itemize}
\end{document}